\documentclass[conference]{IEEEtran}
\usepackage{algorithmicx}
\usepackage[ruled,norelsize]{algorithm2e}
\usepackage{amsmath}
\usepackage{amssymb}
\usepackage{graphicx}
\usepackage{subscript}
\usepackage{multirow}
\usepackage{subcaption}
\usepackage{cleveref}
\makeatletter
\newcommand{\removelatexerror}{\let\@latex@error\@gobble}
\makeatother
\graphicspath{{./}{images/}}

\usepackage{times}
\newtheorem{lemma}{Lemma}
\newenvironment{proof}[1][Proof]{\begin{trivlist}
\item[\hskip \labelsep {\bfseries #1}]}{\end{trivlist}}

\title{Online Anomaly Detection via Class-Imbalance Learning}
\author{Chandresh Kumar Maurya and Durga Toshniwal \\
Indian Institute of Technology, Roorkee\\
Haridwar, Uttarakhand, India \\
ckm.jnu@gmail.com}

\author{\IEEEauthorblockN{Chandresh Kumar Maurya and Durga Toshniwal}
\IEEEauthorblockA{Computer Science \& Engineering\\
Indian Institute of Technology, Roorkee\\
Haridwar-247667, Uttarakhand, India \\
Email: {ckm.jnu, durgatoshniwal}@gmail.com}
\and
\IEEEauthorblockN{Gopalan Vijendran Venkoparao}
\IEEEauthorblockA{Research \& Technology Center\\
RBEI, Bangalore\\
Email: GopalanVijendran.Venkoparao@in.bosch.com }
}
\IEEEoverridecommandlockouts
\IEEEpubid{ \makebox[\columnwidth]{XXX/YY/\$31.00~
\copyright2015
IEEE \hfill} \hspace{\columnsep}\makebox[\columnwidth]{ }}
\begin{document}
\maketitle
\begin{abstract}
 Anomaly detection is an important task in many real world applications such as fraud detection, suspicious activity detection, health care monitoring etc. In this paper, we tackle this problem from supervised learning perspective in online learning setting. We maximize well known \emph{Gmean} metric for class-imbalance learning in online learning framework. Specifically, we show that maximizing \emph{Gmean} is equivalent to minimizing a convex surrogate loss function and based on that we propose novel online learning algorithm for anomaly detection. We then show, by extensive experiments, that  the performance of the proposed algorithm with respect to $sum$ metric is as good as a recently proposed Cost-Sensitive Online Classification(CSOC) algorithm for class-imbalance learning over various benchmarked data sets while keeping running time close to the perception algorithm.  Our another conclusion is that other competitive online algorithms do not perform consistently over data sets of varying size.
 This shows the potential applicability of our proposed approach. 
\end{abstract}
\IEEEpeerreviewmaketitle
\section{Introduction}
Anomaly detection aims to capture behavior in data that do not conform to the normal behavior as expected by domain expert \cite{Chandola2009}. Anomaly detection in online setting is an important task in many real world applications. For example, intrusion detection in computer network, flight navigation system, credit card fraud detection and so on. It is clear that such task require detection of malicious activity on the fly. However, most existing techniques focus on offline training of the model and then use it to detect anomalies \cite{Breunig2000,Goernitz2013,Liang2011,Rose2014}.  

In this paper, we tackle anomaly detection problem from supervised learning perspective in online setting. In supervised learning framework, anomaly detection refers to correctly classifying rare class examples as compared to majority examples. For example, out of 100 persons visiting a doctor for cancer check up, saying a person, who actually have the cancer, not having cancer is more costly than saying he has the cancer when he actually does not. Therefore, we take anomaly detection problem as class-imbalance learning problem. Hence from now and onwards, we will use class imbalance learning to refer to anomaly detection. 

In what follows, we present related work in section 2. Section 3 is devoted to problem formulation in online learning setting and online algorithm for class-imbalance learning. Experimental results are presented in section 4 and finally section 5 concludes the paper. 

\section{Related Work}
Work presented in this paper spans two main themes in data mining and machine learning: Online learning and class-imbalance learning. Although there have been many works in both domain separately \cite{Chawla2002,Chawla2003}, little work has been done that jointly solves online learning and class-imbalance learning. Below we briefly describe work in each domain that closely matches our work. 

\subsection{Online learning}

Online learning aims to process one example at a time. First, it receives an examples and then makes a prediction. If prediction goes wrong, it suffer loss and updates its parameters. \\
Online learning has its origin from classic work of Rosenblatt on perceptron  algorithm \cite{Rosenblatt1958}. Perceptron algorithm is based on the idea of single neuron. It simply takes an input instance and learn a linear predictor of the form $f(\mathbf{w})=\mathbf{w}^T{\mathbf{x}}$, where $\mathbf{w}$ is weight vector and $\mathbf{ x}$ is the input instance. If it makes a wrong prediction, it updates its parameter vector as follows:
\begin{equation}
\mathbf{w}_{t+1}=\mathbf{w_t} +y_t\mathbf{x_t}
\end{equation}
where $\mathbf{w}_{t+1}$ is weight vector at time $t+1$.\\
\cite{Kivinen2010} proposed online learning with kernels. Their algorithm, called $NORMA_\lambda$,  is based on regularized empirical risk minimization which they solved via regularized stochastic gradient descent. They also showed empirically how this can be used in anomaly detection scenario. However, Their algorithm requires tuning of many parameters which is costly for time critical applications. Passive-Aggressive (PA) learning \cite{Crammer2006} is another online learning algorithm based on the idea of maximizing ``margin'' in online learning framework. PA algorithm updates the weight vector whenever ``margin'' is below a certain threshold on the current example. Further, the same author have introduced the idea of slack variable to handle non-linearly separable data. 
\subsection{Class-Imbalance Learning}
Class-imbalance learning aims to correctly classify minority examples. In literature, there exist solutions that are either based on the idea of sampling or weighting scheme. In the former case, either majority examples are under sampled or minority examples are over sampled.  In the latter case, each example is weighted differently and idea is to learn these weights optimally. Some examples of sampling based techniques are SMOTE \cite{Chawla2002}, SMOTEBoost \cite{Chawla2003}, AdaBoost.NC \cite{Shou2010} and so on. Work that used weighting scheme include cost-sensitive learning \cite{Elkan2001},  \cite{Ling2006}, \cite{Liu2006}, \cite{Lo2011} and so on. 

It is worthwhile to mention here that only a few work exists that jointly solves class-imbalance learning and online learning. Below, we mention some work that closely matches our work.
In \cite{Wang2013}, author proposed sampling with online bagging (SOB) for class imbalance detection. Their idea essentially is based on resampling. That is, oversample minority class and under sample majority class from Poisson distribution with average arrival rate of $N/P$ and $R_p$ respectively where $P$ is total number of positive examples, N is total number of negative examples, and $R_p$ is recall on positive examples. \cite{Wang2013} also tries to maximize \emph{Gmean}, but the way they approached to solve the maximization problem is different from our present work. 
\cite{Jialei2014} proposed online cost sensitive classification of imbalanced data. One of their problem formulation is based on maximizing weighted sum of sensitivity and specificity and the other is minimizing weighted cost. Their solution is based on minimizing convex surrogate loss function (modified hinge loss) instead of non-convex 0-1 loss. \cite{Jialei2014} work closely matches our work. But, in section 3 we show the major difference between the two work.

\section{Framework of Class-Imbalance Learning}
\subsection{Problem formulation}
Without loss of generality, consider binary classification problem. Formally, let $\mathcal{X}$ be instance space in $\mathcal{R}^d$ and $\mathcal{Y}$ be label space in $\{-1, +1\}$. We are given samples $S = \{({\bf x}_1,y_1),({\bf x}_2,y_2),...,({\bf x}_T,y_T)\}$ where instance-label pair $( {\bf x}_i,y_i) \in \mathcal{X} \times \mathcal{Y}$, and $i \in \{1,2,\dotsc,T\}$. In online learning setting, no assumption is made about the distribution of the samples and they come sequentially . Let ${\bf x}_t$ be an instance received at time step $t$ and $f_{t}$  be the model that is obtained from previous $t-1$ rounds. Let $\hat{y}$ be the prediction for the $t$-th instance i.e $\hat{y} = sign(f_{t}({\bf x}_t))$,  whereas the value $|f_{t}({\bf x}_t)|$ known as ''margin'', is used as the confidence of the learner on the $t$-th prediction step.  

For binary classification task, let $P$ and $N$ respectively denote the number of positive and negative instances received so far. Let $TP, TN, FP$ and $FN$ denote number of true positive, true negative, false positive and false negative so far.  Mathematically, they are defined as: $TP_t = \{ y =\hat{y}=+1  \}, TN_t =\{y = \hat{y} = -1 \},  FP_t = \{y= -1, \hat{y} = +1\}, FN_t = \{y = +1, \hat{y} = -1\} $ where subscript $t$ denote these metric value at that time step. $TP$ is calculated by summing $TP_t$ from $t=1$ to $T$. Similarly $TN$, $FP$ and $FN$ can be calculated.

Without loss of generality, assume positive instances are minority class. It is well known that maximizing {\it accuracy} as a measure of performance on class-imbalanced data leads to false conclusion. For example, suppose training data contains 100 examples having 95 negative examples and 5 positive examples. If a classifier predicts each example as negative it will have accuracy of $100\%$. Thus missing all the positive examples to classify correctly. 

Therefore, we require a metric that can account for classifying minority class correctly.  $Gmean$ \cite{Kubat1997} is such a metric that evaluates the degree of inductive bias in terms of a ratio of positive accuracy and negative accuracy. Mathematically,  $Gmean$ is defined as:

\begin{equation}
\label{gmean}
Gmean = \sqrt{recall^+ \times recall ^-}
\end{equation}

where $recall^+$ and $recall^-$ denote accuracy  on positive and negative examples respectively and defined as:
\begin{equation*}
recall^+ = \frac{TP}{TP+FN}, recall^- = \frac{TN}{TN + FP}
\end{equation*}•
For simplicity, we assume that $\|\mathbf{x}_t\|\leq 1$ which says that all incoming instances lies within a unit ball. However, this restriction can be relaxed in a more general setting, that is, we can take $\|\mathbf{x}_t\| \leq r$, where $r$ is radious of the Euclidean ball centered at $0$. This condition ensures that cumulative regret, that is, performance of online learner with respect to a fixed learner chosen in hindsight, can be bounded.  Our objective is to maximize \eqref{gmean}. 

Suppose we are given a linear classifier of the form $ f(\mathbf{x}) =\mathbf{w}^T\mathbf{x}$, where $\mathbf{w}$ represents the weight of the linear classifier that we wish to learn in an online manner.   Whenever classifier makes a mistake i.e $yf(\mathbf{x})<0$, we update the weight. How weights are actually updated will be explained in the following section. \vspace{-2ex}

\subsection{Online Setting}
In this subsection, we prove the following lemma due to \cite{Wang2013} for the sake of completeness.  Then we show the hardness of optimizing the equivalent formulation in the lemma and propose to optimize a convex surrogate loss function instead. 
\begin{lemma}
\label{lemma}
Maximizing Gmean as given in \eqref{gmean} is equivalent to minimizing the following objective:
\begin{equation}
\sum\limits_{y_t =+1}\frac{N}{P} I_{(y_t f_t({\bf x_t})<0)} + \sum\limits_{y_t =-1}\frac{P-FN}{P} I_{(y_t f_t({\bf x_t})<0)}
\end{equation}
\end{lemma} \vspace{-2ex}

\begin{proof}
Maximizing \eqref{gmean} is equivalent to maximizing its square. So we can write it as follows:
\begin{equation*}
\begin{split}
Gmean & = \left( {\frac{{TP}}{{TP + FN}}} \right) \times \left( {\frac{{TN}}{{TN + FP}}} \right)\\
& = \left( {\frac{{P - FN}}{P}} \right) \times \left( {\frac{{N - FP}}{N}} \right) \\
&  = \left( {1 - \frac{{FN}}{P}} \right) \times \left( {1 - \frac{{FP}}{N}} \right)\\
&=1 - \left( {\frac{{FN}}{P} + \frac{{FP}}{N} - \frac{{FN.FP}}{{P.N}}} \right)
\end{split}
\end{equation*}
where in step 2, we used the fact that $P=TP+FN$ and $N=FP+TN$. 
Hence we minimize:
\begin{equation}
\begin{split}
Gmean &=\frac{{FN}}{P} + \frac{{FP}}{N} - \frac{{(FN)(FP)}}{{PN}}\\
&  = \left( {\frac{N}{P}} \right)FN + \left( {\frac{{P - FN}}{P}} \right)FP\\
&  = \sum\limits_{{y_t} =  + 1} {\frac{N}{P}} {I_{({y_t}  f_t({\bf x_t}) < 0)}} + \\ &\qquad
\sum\limits_{{y_t} =  - 1} {\frac{{P - FN}}{P}} {I_{({y_t} f_t({\bf x_t}) < 0)}}
\end{split}
\end{equation}•
\end{proof}
where in step 2, we have taken off the constant $N$ from the equation  \footnote{It can be shown that problem formulation presented in this paper is similar to one presented in \cite{Jialei2014} but not equivalent since they were minimizing weighted sum of sensitivity and specificity where weights are decided by Laplace estimation. In our formulation, we have directly estimated these weights in terms of $P, N $, and $FN$.} without changing the optima and $I(C)$ is an indicator function that outputs 1 when argument $C$ is true and $0$ otherwise. \qquad\qquad \qquad\qquad\qquad\qquad\qquad\qquad\qquad$\square$.

Lemma \ref{lemma} gives an alternative objective to maximize \emph{Gmean}. But, it involves indicator function that is non-convex. Hence, we resort to convex relaxation techniques. More specifically, we use the following convex surrogate loss function:
\small
\begin{equation}
\label{loss}
\mathcal{L}(\mathbf{w};(\mathbf{x},y))=\max \left( {0,  \rho- y f_t({\bf x_t}) } \right) 
\end{equation} 
\normalsize
where
\begin{equation*}
\rho = \left( \left( {\frac{N}{P}} \right){I_{(y = 1)}} + \left( {\frac{{P - FN}}{P}} \right){I_{(y =  - 1)}}\right)
\end{equation*}•

We see that (\ref{loss}) is similar to modified hinge loss function.

Our next goal is to put the modified loss function into online learning framework. In online  learning, learner receives one example at a time and makes it prediction. If it goes wrong, it updates it prediction function $f$. Here we will cast this problem into empirical risk minimization(ERM) learning framework. In offline learning under ERM framework, measure of quality of $f$ is {\it expected risk} \cite{Kivinen2010}

\begin{equation}
R_{emp}[f,S] = E_{(x,y)\sim \mathcal{P}}  [(\hat{y}),y] 
\end{equation}

Since distribution $\mathcal{P}$ is unknown in general, one instead minimizes empirical risk
\begin{equation}
R_{emp}[f,S]= \frac{1}{T}\sum_{t=1}^{T} \mathcal{L}(f(x_t),y_t)
\end{equation}
One can further regularizes empirical risk to avoid overfitting.
\begin{equation}
R_{emp}[f,S]= \frac{1}{T}\sum_{t=1}^{T} \mathcal{L}(f(x_t),y_t) + \frac{\lambda}{2}||f||^2
\end{equation}
where $\lambda$ is a regularization parameter that controls trade-off between complexity of the model and correctness of the prediction.

Since we are interested in online learning, we define an instantaneous risk \cite{Kivinen2010} of using prediction function $f_t$  on example $(x_t,y_t)$ as follows:
\begin{equation}
\label{instantrisk}
R_{emp}[f_t;(x_t,y_t)]=  \mathcal{L}(f_t(x_t),y_t) +\frac{\lambda}{2}||f_t||^2
\end{equation}

Next, we need to determine how our model is performing over sequence of inputs during online learning. For that, cumulative mistake is often used and defined as:
\begin{equation}
\mathcal{L}_{CUM}[f_t,T]= \sum_{t=1}^T \mathcal{L}(f_t(\mathbf{x_t}),y_t)
\end{equation}
We note here that $f_t$ is tested on an example $(\mathbf{x_t},y_t)$ which was not used for fitting $f_t$. Thus, if we can guarantee low cumulative mistake, we can prevent overfitting without the need of regularizer \cite{Kivinen2010}. We used this fact in our current work and do not provide explicit regularization parameter.

Thus our objective is to minimize (\ref{instantrisk}). To this end, we use online gradient descent \cite{Zinkevich2003} to optimize (\ref{instantrisk}).

\begin{equation}
\mathbf{w_{t+1} = \mathbf{w_t} - \tau \nabla \mathcal{L}_{t}(w_t) }
\end{equation}
where $\nabla \mathcal{L}_{t}$ is loss incurred at time $t$ on example $\mathbf{x}_t$.
To find $\nabla \mathcal{L}_{t}$ for our loss function (\ref{loss}), we differentiate it with respect to $\mathbf{w}_t$ and update rules becomes:

\begin{equation}
\label{update}
\mathbf{w}_{t+1}  =
 \left \{
  		 \begin{array}{ll}
			\mathbf{w}_{t} +\tau y_{t}\mathbf{x}_{t} & if\quad \mathcal{L}_{t}(\mathbf{w}_{t}) >0 \\
			\mathbf{w}_{t} &  \quad otherwise
       	\end{array}
\right.
\end{equation}

\subsection{Algorithm}
Our proposed algorithm Online G-MEAN (OGMEAN) is given in Algorithm \ref{ogmean}. As we can see that OGMEAN requires only one parameter to tune; $\tau$, the learning rate. In general it is set to $1/\sqrt{t}$. However, for simplicity we set it to a constant in our experiments.\\
In is clear that OGMEAN takes time proportional to $O(T\times n)$, which is linear in the number of dimension of the input instances as well as number of received instances so far. 
%\HRule
%{\bf Algorithm 1.} Online G-mean(OGMEAN) algorithm 
% \hrule
%\begin{algorithm}
%\label{ogmean}
%\SetKwInOut{Input}{Input}\SetKwInOut{Output}{Output}\SetKwInOut{Initialization}{Initialization}
% \Input{Learning rate $\tau$}
% \Initialization{$\mathbf{w_1} =\mathbf{0}$}
% \Output{Weight vector $\mathbf{w}_{T+1}$}
% 
% \For{ $t=1,\dotsc,T$}{
%   receive instance: $x_t$\;
%   predict :$\hat{y}_t$ = sign($\mathbf{w_t . x_t})$\;
%   receive correct label: $y_t \in \{-1,+1\}$ \;
%   suffer loss: $\mathcal{L}_t(\mathbf{w_t}) $ as given in (\ref{loss})\;
%   \eIf{$\mathcal{L}_t(\mathbf{w_t}) >0$}{
%$ \mathbf{w}_{t+1}  =\mathbf{w}_{t} +\tau y_{t}x_{t}$ \;
%   
%   }{
%   $\mathbf{w}_{t+1}  =\mathbf{w}_{t}$ \;
%  }
% }
% \hrule
% %\caption{How to write algorithms}
%\end{algorithm}

\begin{figure}[!t]
 \removelatexerror
  \begin{algorithm}[H]  
   \caption{Online G-mean(OGMEAN) algorithm }
\label{ogmean}
%   \SetKwInOut{Input}{Input}\SetKwInOut{Output}{Output}\SetKwInOut{Initialization}{Initialization}
 {\bf Input}: Learning rate $\tau$ \\
 {\bf Initialization} $\mathbf{w_1} =\mathbf{0}$ \\
 {\bf Output}: Weight vector $\mathbf{w}_{T+1}$\\
 
 \For{ $t=1,\dotsc,T$}{
   receive instance: $x_t$\;
   predict :$\hat{y}_t$ = sign($\mathbf{w_t . x_t})$\;
   receive correct label: $y_t \in \{-1,+1\}$ \;
   suffer loss: $\mathcal{L}_t(\mathbf{w_t}) $ as given in (\ref{loss})\;
   \eIf{$\mathcal{L}_t(\mathbf{w_t}) >0$}{
$ \mathbf{w}_{t+1}  =\mathbf{w}_{t} +\tau y_{t}x_{t}$ \;   
   }{
   $\mathbf{w}_{t+1}  =\mathbf{w}_{t}$ \;
  }
 }

 %\caption{How to write algorithms}
\end{algorithm}
\end{figure}
\subsection{Relative Loss Bound for OGMEAN }
Following \cite{Jialei2014}, we can bound the regret of OGMEAN algorithm.  Below we just state the lemma without proof. 

\begin{lemma}
\label{mistake}
Let $S=\{(\mathbf{x_t},y_t)\}_{t=1,\dotsc,T}$ be the sequence of $T$ examples where $\mathbf{x_t} \in \mathcal{X}$, $y_t \in \mathcal{Y}$ and $||x_t|| \leq 1$ for all $t$. Then for any $\mathbf{w} \in \mathcal{X}$, by setting $\tau = ||\mathbf{w}||\sqrt{T}$, the following holds for OGMEAN:

\begin{equation*}
\sum_{t=1}^T \mathcal{L}_t (\mathbf{w}_t) \leq \sum_{t=1}^T \mathcal{L}_t (\mathbf{w}) + ||\mathbf{w}|| \sqrt{T}         \;\;\; \;\;\; \;\;\; \;\;\;\;\;\;\;\;\;\;\;\;\;\;    \square.
\end{equation*}

\end{lemma}

\section{Experiments}
In this section, we experimentally validated the accuracy of the proposed OGMEAN algorithm over various bench marked data set which can be freely downloaded from LIBSVM website\footnote{http://www.csie.ntu.edu.tw/~cjlin/libsvmtools/datasets}. 
A brief summary of the data set and class-imbalance ratio is given in Table \ref{data}. All the algorithms were run in MATLAB 2012a (64 bit version) on 64 bit Windows 8.1 machine.
\begin{table}
\centering
\caption{Summary of datasets used in the paper}
\label{data}
\begin{tabular}{|l|c|c|l|}
\hline
Dataset & \# Examples & \#Features &\#Pos:Neg \\ \hline \hline
covtype & 581012 & 54 & 1:1 \\\hline
german & 1000 & 24& 1:2.3 \\\hline
svmguide3 & 1243&21&1:3\\\hline
ijcnn1 & 141691&22 & 1:10.44 \\\hline
\end{tabular}•
\end{table}
We compare our algorithm with recently proposed $CSOC_{sum}$ algorithm \cite{Jialei2014} with respect to a metric called {\it sum} which is defined as:
\begin{equation}
\label{sum}
sum = n_p\times sensitivity+n_n \times specificity
\end{equation}•
where $n_p$ and $n_n$ are weight parameter which in  \cite{Jialei2014} are set manually to 0.5 each and sensitivity and specificity are the same as recall on positive and negative examples respectively.
It is claimed in \cite{Jialei2014} that  $CSOC_{sum}$ algorithm beats state-of-the-art online algorithms for class-imbalance problem with respect to {\it sum} metric. The algorithms compared  in \cite{Jialei2014} are Perceptron \cite{Rosenblatt1958}, ROMMA, agg-ROMMA, PA-I, PA-II, CPA\textsubscript{PB} \cite{Crammer2006} and PAUM \cite{Li2002}. Since $CSOC_{sum}$ performs equal or better than all the above algorithms, we only compare OGMEAN to $CSOC_{sum}$ with respect to {\it sum} metric in this paper. We also show the mistake rate, number of updates and running time of all the algorithms for fair comparison. For this purpose, we used the LIBOL online learning library \footnote{http://stevenhoi.org}.
\subsection{Evaluation of Weighted Sum Performance}

\begin{figure*}
\centering
\begin{subfigure}{.33\textwidth}
  \centering
 \includegraphics[width=2in, height=2in]{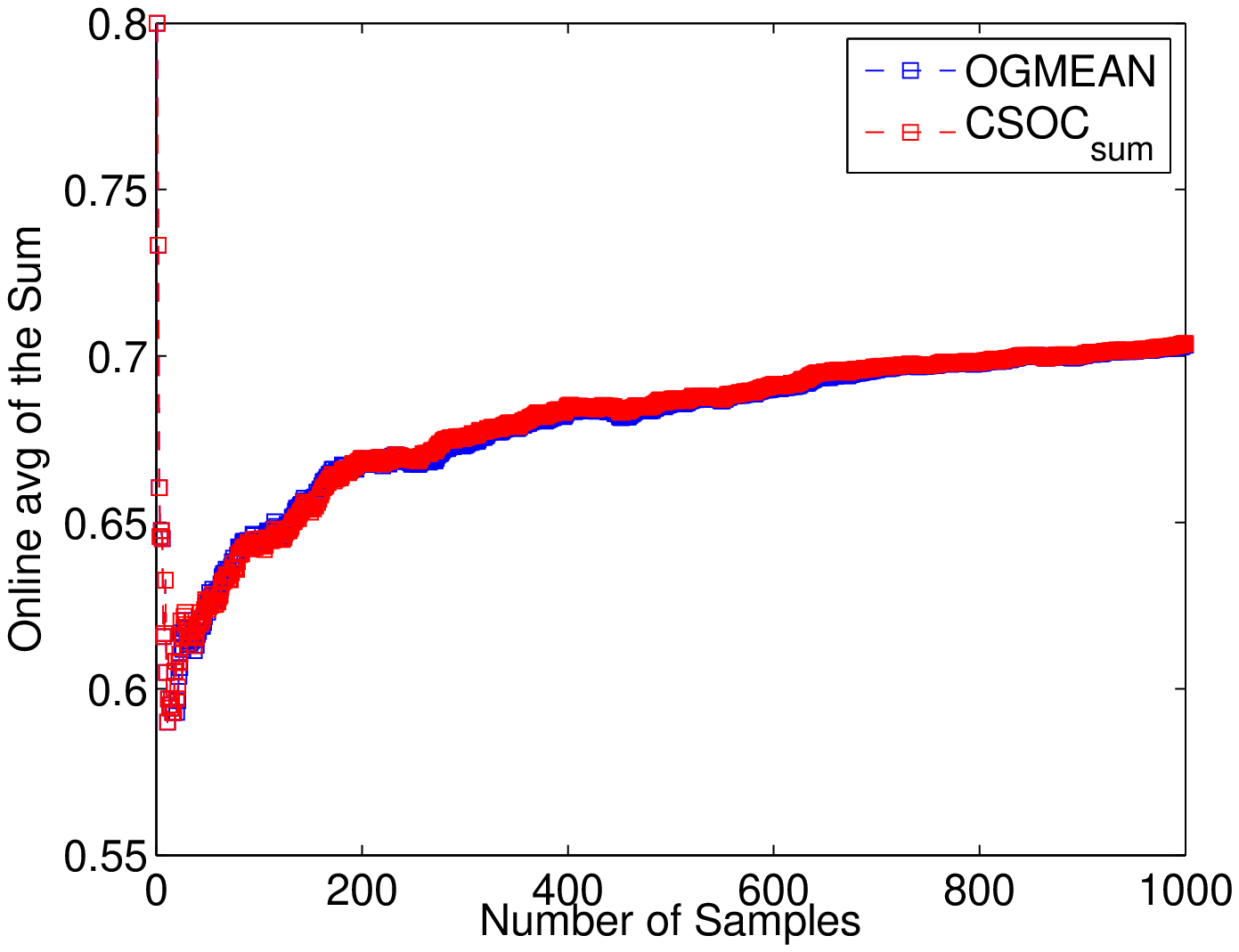}
\caption{german}
\label{german}
\end{subfigure}%
\begin{subfigure}{.33\textwidth}
  \centering
 \includegraphics[width=2in, height=2in]{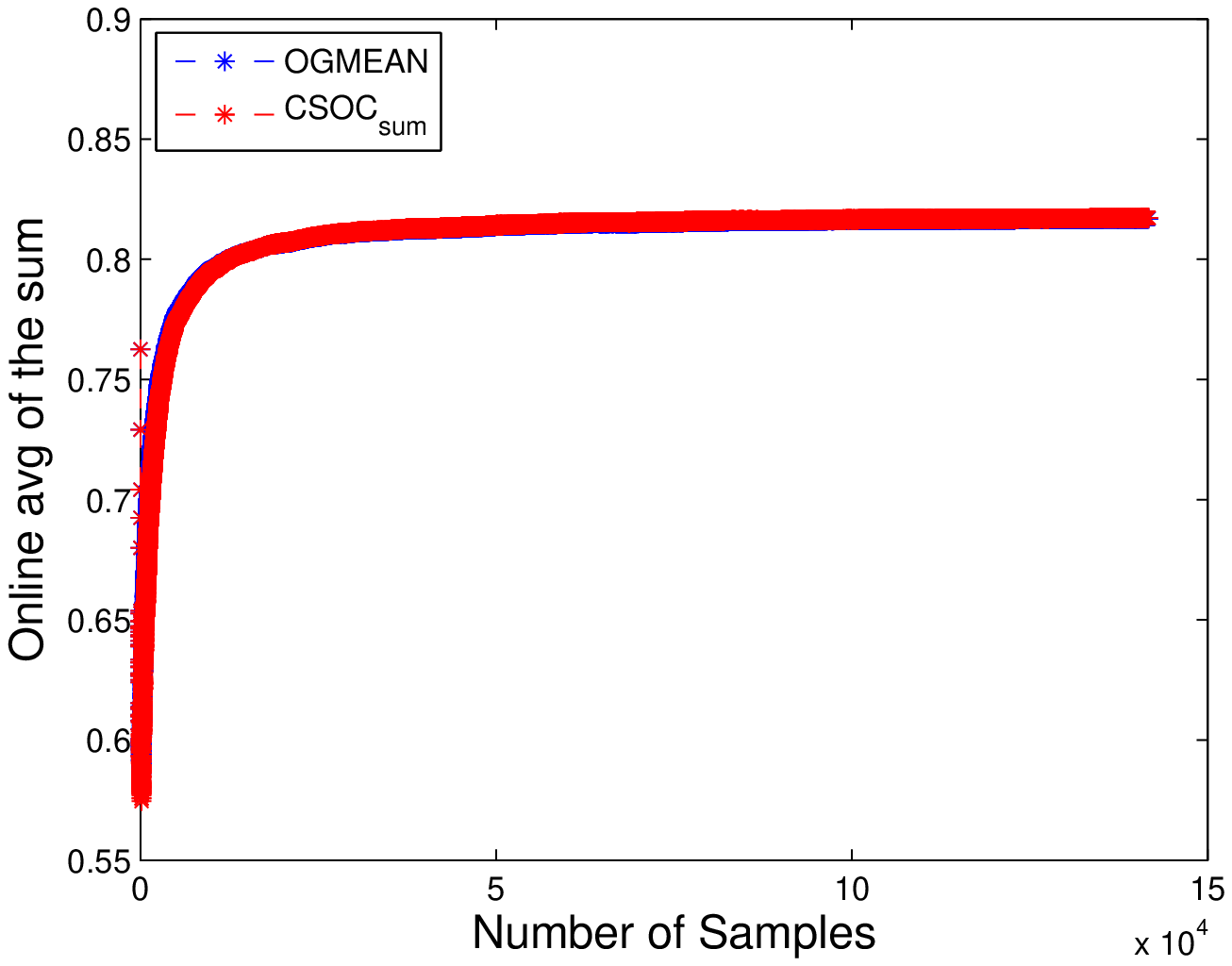}
\caption{ijcnn1}
\label{ijcnn}
\end{subfigure}
\begin{subfigure}{.33\textwidth}
  \centering
\includegraphics[width=2in, height=2in]{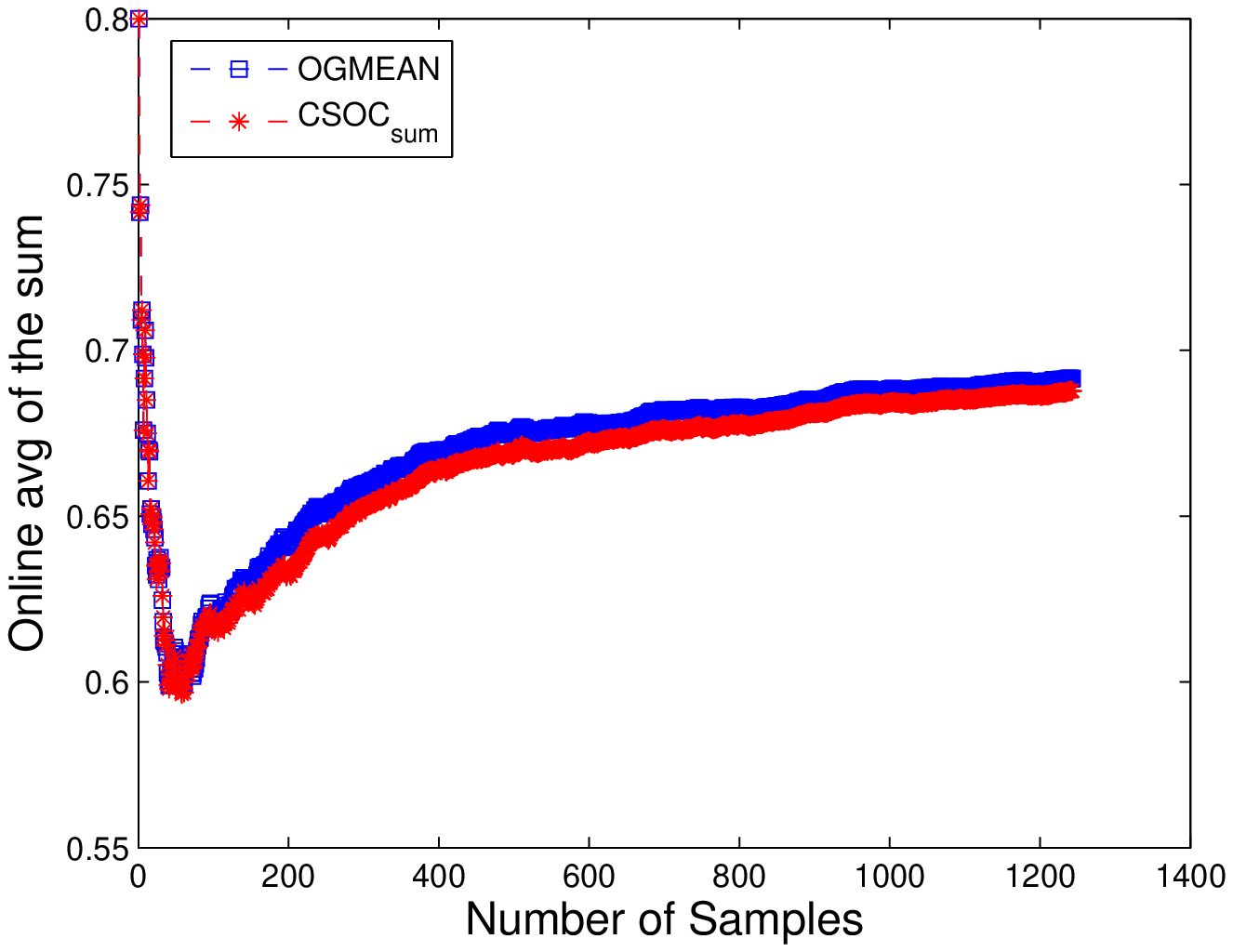}
\caption{svmguide3}
\label{svmguide3}
\end{subfigure}
\caption{Online ``sum' 'Performance of OGMEAN on various data sets }
\end{figure*}
\begin{figure}
\centering
\includegraphics[width=3.35in, height=2in]{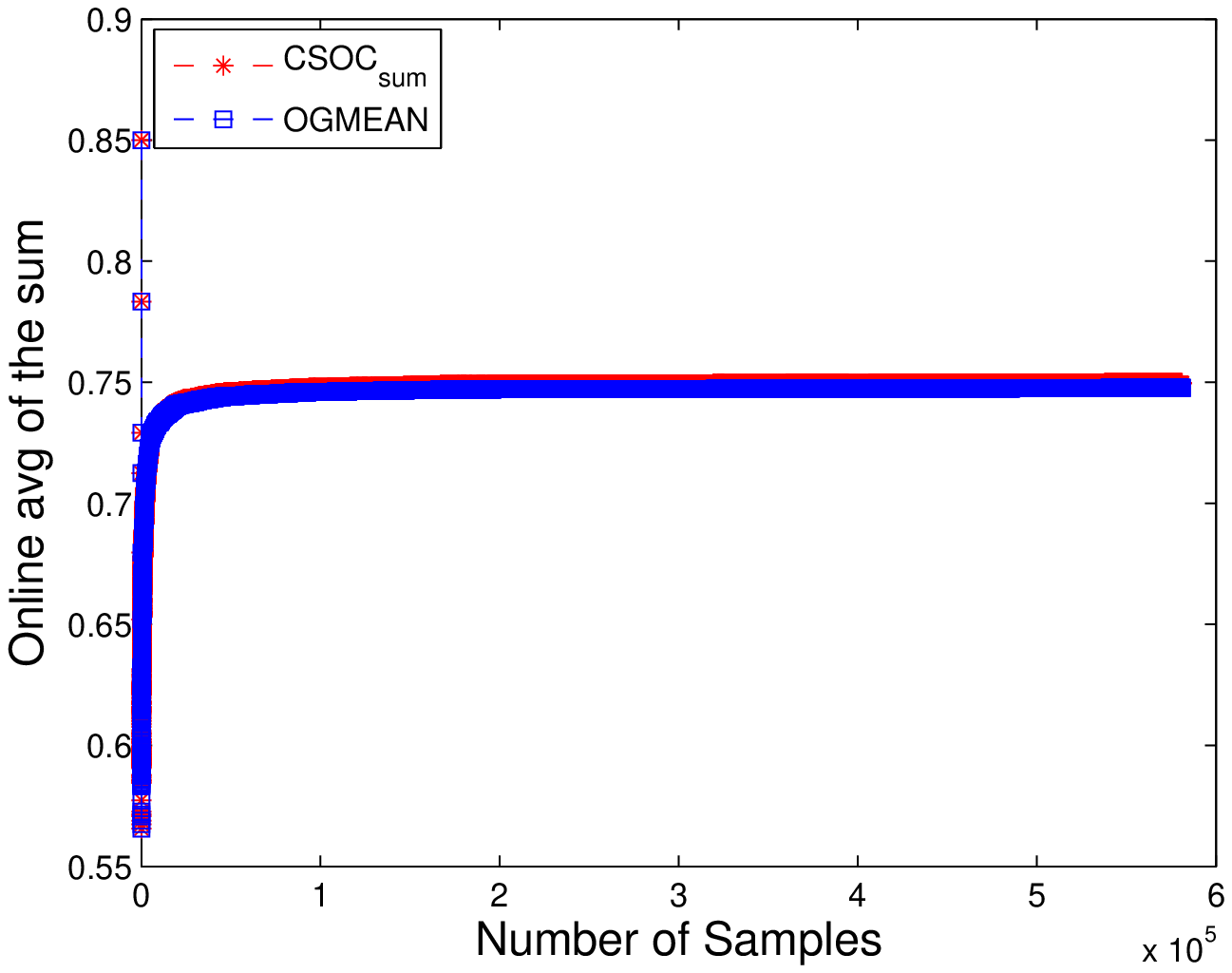}
\caption{Online ``sum'' performance on ``covtype'' dataset}
\label{covtype}
\end{figure}

We conducted a comparative study  of CSOC\textsubscript{sum} and OGMEAN algorithms where we set learning rate parameter $\tau$ equals to 0.2  and weights for ``sum'' i.e $n_p$ and $n_n$ equal to 0.5 over all the data sets and both algorithms as done in CSOC\textsubscript{sum}.   Online ``sum'' performance of the two algorithms are shown in  \Cref{german,ijcnn,svmguide3}, \ref{covtype} and Table \ref{compcsocgmean} over four data sets: ``german'', ``ijcnn1'',``svmguide3'', and ``covtype''. We can draw several conclusions from these results. First, we can see in \Cref{german,ijcnn,svmguide3} that OGMEAN achieves equal or higher ``sum'' value as compared to CSOC\textsubscript{sum} over all the data sets. It shows the potential drawback in using Laplace estimation to estimate $\rho$ in CSOC\textsubscript{sum} algorithm. 
From the Table \ref{compcsocgmean}, we infer that OGMEAN beats CSOC\textsubscript{sum} algorithm in terms of Mistake rate, No of updates applied on weights and CPU time on almost all the data sets. It again justifies the potential benefits of using OGMEAN in real world applications. 

\subsection{Comparative Study}
\begin{figure*}
\centering
\includegraphics[width=7.5in, height=3in]{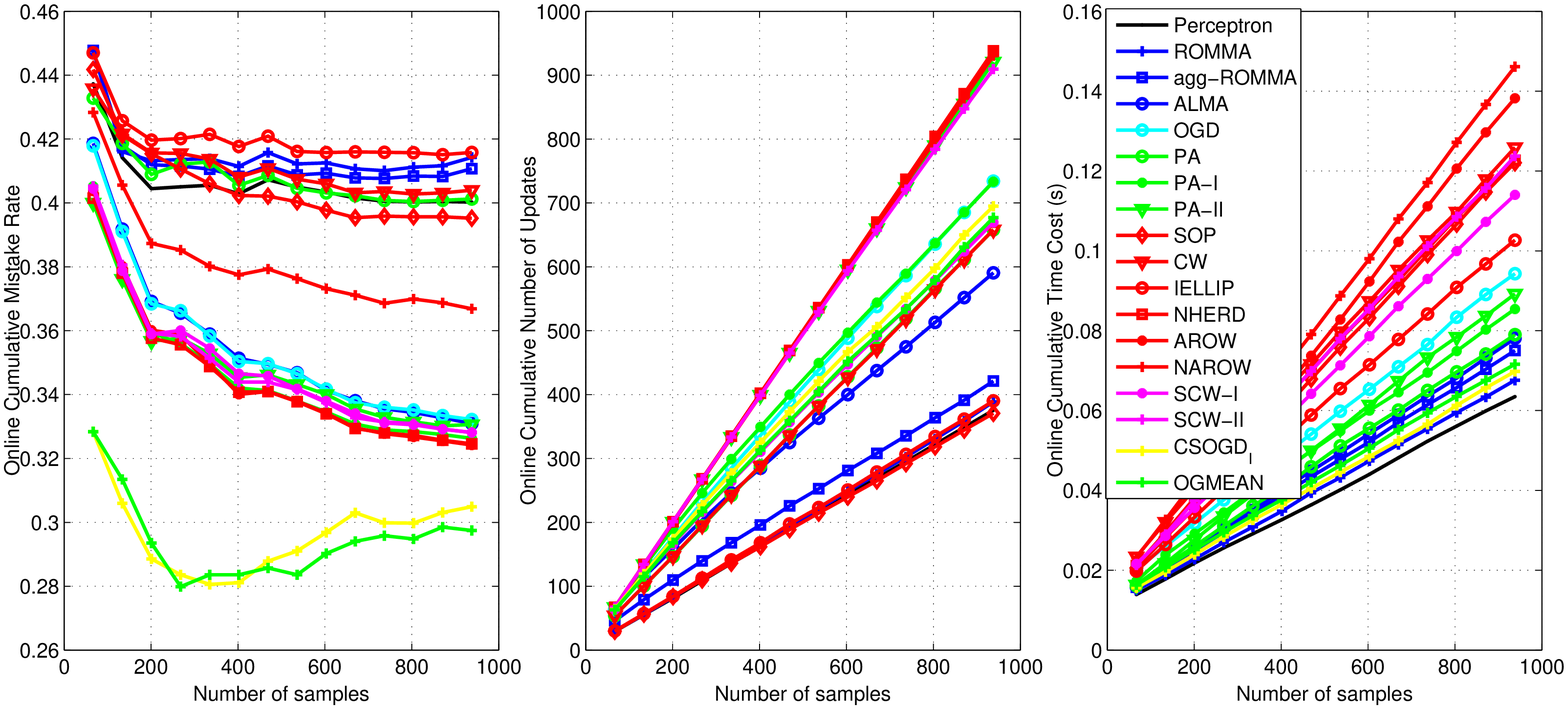}
\caption{Comparative study of mistake rate, cumulative number of updates and cumulative time cost over $german$ data set}
\label{comp1}
\end{figure*}

\begin{figure*}
\centering
\includegraphics[width=7in, height=3in]{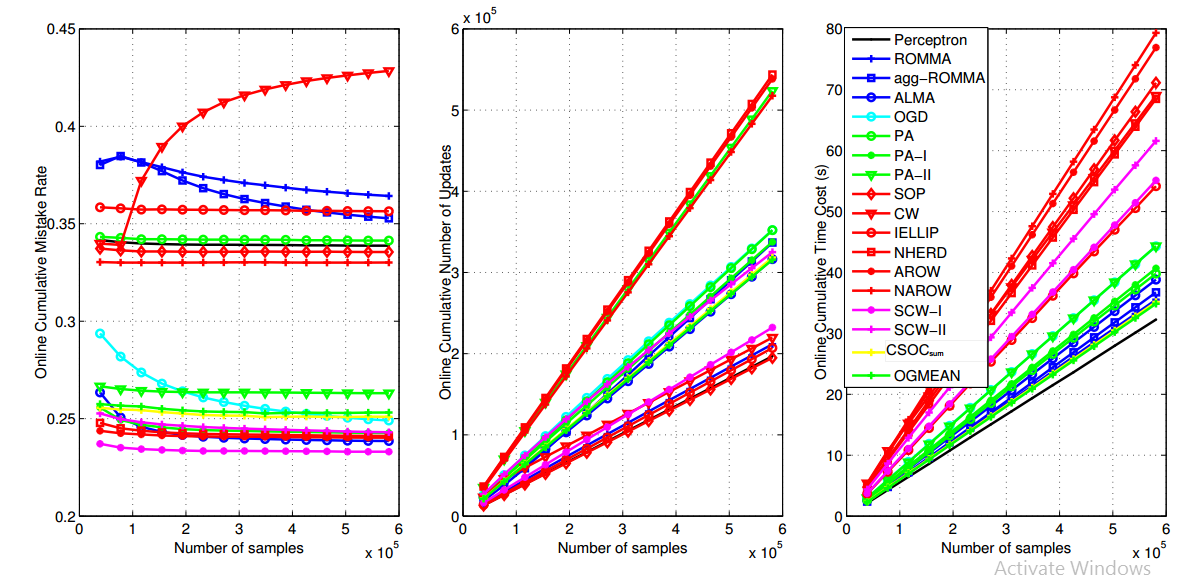}
\caption{Comparative study of mistake rate, cumulative number of updates and cumulative time cost over $covtype$ data set}
\label{comp2}
\end{figure*}

We conducted another experiment for comparing mistake rate, cumulative number of updates and cumulative time cost over ``covtype'' and ``german3'' data sets. For all the algorithms compared, cost parameter $C$ is set to 1 except ALMA for which the value of $C$ is $\sqrt{2}$. We kept the values of all other parameters same as given in LIBOL implementation.  From \Cref{comp1}, we observe that both CSOC\textsubscript{sum} and OGMEAN   has cumulative running time approaching perception algorithm. In terms of the number of mistakes, both CSOC\textsubscript{sum} and OGMEAN outperform all other algorithms. On the other hand, in \Cref{comp2}, we observe that both CSOC\textsubscript{sum} and OGMEAN did not do well in comparison to SCW-I, SCW-II, NHERD, and ALMA in terms of the number of mistakes. However, CSOC\textsubscript{sum} and OGMEAN took less time as compared to all other algorithms except PERCEPTRON. 

In addition, we also tested comparative performance of all the algorithms over other bench marked data sets (``ijcnn1'', ``svmguide3'') where we found that performance of CW, SCW-I, and SCW-II was better than all other algorithms. However, result presented in this paper indicates that aforementioned algorithms are not so well performing on ``german'' and ``covtype'' as compared to CSOC\textsubscript{sum} and OGMEAN.  This leads us to a conclusion that there is lack of consistency of performance in terms of the number of mistakes made by these algorithms except CSOC\textsubscript{sum} and OGMEAN.  

\section{Conclusion}
In the present work, we tackled  binary class imbalance learning under online learning framework. We maximize popular \emph{Gmean} metric for class imbalance problem. We showed that maximizing \emph{Gmean} equivalent formulation is non-convex and hence used convex surrogate loss function under empirical risk minimization framework. We compared our OGMEAN algorithm performance with recently proposed CSOC\textsubscript{sum} algorithm over various bench marked data sets. It is found that directly optimizing weighted sum of sensitivity and specificity where weights are learned using Laplace estimation techniques is less efficient as compared to directly optimizing equivalent formulation of maximizing \emph{Gmean}. We also showed mistake rate, cumulative time cost and number of updates of our algorithm with respect to many other online learning algorithms and concluded that its performance is as good as or better than these recent online algorithms. 

In our future work, we plan to extend the work to multi-class setting. Concretely, how can we optimize \emph{Gmean} metric for multi-class in online scenario? The problem is that \emph{Gmean} for multi-class is non-decomposable loss function that prohibits us to use any existing optimization techniques. So, further work is required in this direction. 

  \section*{Acknowledgment}
 This research work is supported by the Prime Minister’s Fellowship for Doctoral Research, joint initiative by  Confederation of Indian Industry (CII) and  industry partner Robert Bosch.
\begin{table*}
\renewcommand{\arraystretch}{1.3}% for space bt rows
\centering
\caption{Evaluation of performance of CSOC\textsubscript{sum} and OGMEAN algorithms with respect to ``sum'' metric } \smallskip
\label{compcsocgmean}
\begin{tabular}{|c|c|c|c|c|}% use p for text wrap in column
\hline 
Data Set & Algorithm 	&Mistake rate	&No. of updates  & CPU time(sec) \\ \hline

\multirow{2}{*}{covtype} & CSOC\textsubscript{sum}  &  0.2511 $\pm$ 0.0003	&  318603.35 $\pm$ 116.12 &	 34.4843 $\pm$ 1.0240   \\ \cline{2-5}
& OGMEAN  &  0.2534 $\pm$ 0.0003	& {\bf 315907.00 $\pm$ 125.20} &  {\bf	 33.9553 $\pm$ 0.2790 } \\ \cline{1-5}

\multirow{2}{*}{german}   & CSOC\textsubscript{sum}&  0.3235 $\pm$ 0.0087	&  739.15 $\pm$ 5.81 	& 0.0603 $\pm$ 0.0036   \\\cline{2-5}

& OGMEAN& {\bf 0.3231 $\pm$0.0082}	&  {\bf 716.25 $\pm$ 4.84} & {\bf  0.0599 $\pm$ 0.0037 }\\\cline{1-5}
\multirow{2}{*}{svmguide3}  &CSOC\textsubscript{sum} &  0.2800 $\pm$ 0.0053	 & 857.65 $\pm$ 8.16 &	 0.0722 +/- 0.0046 \\\cline{2-5}

&OGMEAN & {\bf0.2750 $\pm$ 0.0077}	& {\bf  787.90 $\pm$ 9.34} &	{\bf  0.0724 $\pm$ 0.0041} \\\cline{1-5}

\multirow{2}{*}{ijcnn1}   &CSOC\textsubscript{sum} &   0.2742 $\pm$ 0.0007	 & 82706.60 $\pm$ 119.07& 	 7.8521 +/- 0.1987\\\cline{2-5}

&OGMEAN &  0.2754 $\pm$ 0.0008	&  {\bf 81877.85 $\pm$ 102.83} &	{\bf  7.8316 $\pm$ 0.0529 }
\\\hline
\end{tabular}
\end{table*}
%\clearpage 
\bibliographystyle{IEEEtran}
\bibliography{./ic3}

\end{document}